\documentclass{article}

\usepackage[english]{babel}
\usepackage{csquotes}
\usepackage[letterpaper,top=2cm,bottom=2cm,left=3cm,right=3cm,marginparwidth=1.75cm]{geometry}

\usepackage{amsmath}
\usepackage{graphicx}
\usepackage[colorlinks=true, allcolors=blue]{hyperref}
\usepackage[authordate,backend=biber]{biblatex-chicago}
\begin{document}
\title{Applying Machine Learning and AI Explanations
\\to Analyze Vaccine Hesitancy}
\author{Jian Lange\thanks{
%Jian Lange, MBA, 
Principal Product Manager, Environmental Systems Research Institute (Esri), 380 New York Street, Redlands, CA, email: jlange@esri.com}
\and \addtocounter{footnote}{2}
Carsten Lange\thanks{
%Carsten Lange 
%(corresponding author), Ph.D., Professor at 
Department of Economics, California State Polytechnic University, Pomona, 3801 West Temple Avenue
Pomona, California 91768 email: clange@cpp.edu }}

\maketitle 
\begin{abstract}
The paper identifies and quantifies the impact of race, poverty, politics and age on COVID-19 vaccination rates in counties across the continental US. Both traditional Ordinary Least Square (OLS) regression analysis and Random Forest machine learning algorithms are applied to quantify contributing factors for county level vaccination hesitancy. With the machine learning model, joint effects of multiple variables (race/ethnicity, partisanship, age etc.) are considered simultaneously to capture the unique combination of what factors affect the vaccination rate. By implementing a state-of-the-art Artificial Intelligence Explanations (AIX) algorithm, it is possible to solve the black box problem with machine learning models and provide answers to the “how much” question for each measured impact factor in every county. 

For most counties a higher percentage vote for Republicans, a greater African American population share, and a higher poverty rate lower the vaccination rate. While  a higher Asian population share increases the predicted vaccination rate. The impact on the vaccination rate from the Hispanic population proportion is positive in the OLS model, but only positive for counties with very high Hispanic population (65\% and more) in the Random Forest model. Both the proportion of seniors and the one for young people in a county have a significant impact in the OLS model --- positive and negative, respectively. In contrast, the impacts are ambiguous in the Random Forest model. 

Because results vary between geographies and since the  AIX algorithm is able to quantify vaccine impacts individually for each county, this research can be tailored to local communities. This way it is a helpful tool for local health officials and other policy makers to improve vaccination rates. 

An interactive online mapping dashboard that identifies impact factors for individual U.S. counties is available at \href{https://www.cpp.edu/~clange/vacmap.html}{https://www.cpp.edu/\texttildelow clange/vacmap.html}. It is apparent that the influence of impact factors is not universally the same across different geographies.

\vspace*{1cm}

%\textbf{Keywords:} "COVID-19", "Vaccine Hesitancy", "Geographical Information %Systems (GIS)",  "Machine Learning","Random Forest"
\end{abstract}

\newpage
\section{Introduction}

The U.S. was one of the first countries to roll out the COVID-19 vaccines. However, vaccination doses administered per 100 people in the U.S. are among the lowest in the developed world (\cite{NytHolde2021}). Vaccine hesitancy is not a new phenomenon in the US and it has been suggested that age, political affiliation, and income may all play a role. In addition, racial disparities had also been observed as a possible factor. In order to assist policy makers to develop better strategies to decrease hesitance and increase access, it is critical to identify and quantify contributing factors associated with COVID-19 vaccination rates in the US, especially now in  winter with more indoor activities approaching and new COVID variants spreading.

This paper applies a multivariate Ordinary Least Square (OLS) regression analysis as well as a machine learning regression (Random Forest) to quantify contributing factors for county level vaccination hesitancy in the continental US. Joint effects of multiple variables (race/ethnicity, politics, age etc.) are considered simultaneously to capture and quantify which  factors affect the vaccination rate. By implementing a state-of-the-art Artificial Intelligence Explanations (AIX) algorithm, it is possible to solve the black box problem that is often related with machine learning models and provide answers to the “how much” question for each measured impact factor in every individual county. 

An interactive GIS online dashboard is made available so that one can search and select a county in the continental U.S., display a diagram to show the value of each contributing factor, and see by how much each factor in the selected county impacts the predicted vaccination rate.\footnote{See \href{https://www.cpp.edu/~clange/vacmap.html}{https://www.cpp.edu/\texttildelow clange/vacmap.html} for an interactive map.} This impact is also known as the  SHAP value of the predictor variable.\footnote{SHAP (SHapley Additive exPlanations) is an AIX tool. Based on a trained machine learning model SHAP values quantify for each observation (county) the impact that a predictor variable such as political affiliation, race/ethnicity, age, or income has on the prediction (vaccination rate).} It is apparent that the influence of a variable is not universally the same across different geographies (counties). Consequently, county specific strategies need to be applied to help increasing vaccination rates. 

By highlighting the variable impact of different factors in each county, this research can be a helpful tool for local health officials and others to find strategies to improve vaccination rates in individual communities that are currently under-vaccinated. 

The paper is structured as follows: Section \ref{sec:methodology} describes the data sources and the methodology used for the research. In Section \ref{SecAnalysis} we use a traditional linear OLS regression  model (see Section \ref{SubSecRegression})  and a Random Forest machine learning model (see Section \ref{SecCountyRandFor}) to predict the vaccination rate for continental U.S. counties. The Random Forest model allows to consider non-linear impacts as well as interaction effects between variables. The idea behind a Random Forest Model is presented in Section \ref{SecCountyRandForIdea}, the prediction results are discussed in Section \ref{SecCountyRandForPrediction}, and the impact factors for vaccination hesitancy for the Random Forest Model  are analyzed for each county separately in Section \ref{SecCountyShap} using SHAP values.
In Section \ref{SecKeyFindings} we visualize the SHAP values and derive some important key findings. In Section \ref{SecShapTrends} we analyze for each predictor variable the 
direction and trend of the impact on the predicted vaccination rate by generating and plotting SHAP values for all counties. In Section \ref{SecUniqueInsights} we compare the SHAP values for some selected counties. This will give us a unique insights into the vaccination behavior of individual counties and will show that different vaccination strategies need to be implemented for different counties.

\newpage
\section{Methodology and Data}
\subsection{Methodology \label{sec:methodology}}

The goal is to identify and quantify impact factors that can be used to explain different COVID-19 vaccination rates in the counties of the United States.
%and the ZIP code areas in California. 
Impact factors for vaccination rates are analysed on the national level based on county data. 
%and for the State of California based on Zip-Code data . ZIP codes level data where not %available for most other states.

%For each jurisdiction 
As a first approach we apply an Ordinary Least Square (OLS) model (see Sections \ref{SubSecRegression})
%and \ref{SecZipOls}) 
to identify statistically significant predictor variables based on hypothesis tests. In Sections \ref{SecCountyRandFor} 
%and \ref{SecZipRandFor} 
we use a Random Forest machine learning model.\footnote{Random Forest was introduced by \cite{Breiman2001}.} The Random Forest machine learning improves the predictive quality. However, machine learning models are often treated as black-boxes without the capacity to  quantify the impacts of specific predictor variables for the final prediction. More recently, a new methodology (SHAP values) was introduced into the machine learning literature to quantify the impact of predictor variables on the estimated value.\footnote{The  basic idea behind SHAP values goes back to \cite{Shap54} who introduced it as a game theory approach. \cite{Lundberg2018} modified the methodology to interpret machine learning results and developed a related R package for tree based models such as Random Forest.} We apply SHAP value analysis to the results of the Random Forest model and will be able to determine by how much each of our (county) predictions is determined by each of the predictor variables.\footnote{Mazzanti provides a very  intuitive introduction to SHAP values (see \cite{Mazzanti2020Shap}).}

%For ZIP Code areas we measured the vaccination rate as the rate of adult persons at least %once vaccinated over the number of all adults in the ZIP code area. We believe this rate %best reflects how well the relevant geographical area is protected against COVID-19. %Unfortunately, this metric was not available on the national level for U.S. counties. %Therefore, we used the number of fully vaccinated adult persons instead.

As the dependent variable we use the rate of Fully Vaccinated Adult Persons. We choose this variable over the variable Adult Persons who at Least Received One Shot, because the latter consist of three very diverse groups with different vaccination behaviors: i) individuals, who already completed their vaccination cycle, ii) individuals who received one shot but are not yet eligible for the second shot, and iii) individuals who decided not to complete the vaccination cycle.

Based on a correlation analysis of several variables, we chose the following predictor variables:
\begin{itemize}
\item Race/Ethnicity: proportion of African Americans, Asian Americans, and Hispanics for the relevant geography.
\item Political Affiliation: Number of voters who voted for the Republican presidential candidate as a proportion of the sum of voters who voted for the Democratic or Republican presidential candidate in the 2020 presidential election.
\item Age Groups: Proportion of young adults (20-25 years) and Proportion of older adults (65 years and older).
\item Income related: To control for income effects  we used the proportion of households receiving food stamps in the related geography as a predictor variable.\footnote{We also considered Median Household income and the Social Vulnerability Index as defined be the Center for Disease Control (CDC). The results were similar to the proportion of food stamp recipients but less significant. A possible reason could be that the proportion of food stamp variable better represents low income.}
\end{itemize}

For both the OLS and the Random Forest regressions the observations are weighted with the population of the the related geography. This avoids that a 
%ZIP Code or 
county area with a small population has the same weight in the analysis as 
%a Zip Code or 
a county area with a high population. This is important because population varies drastically from county to county in the U.S.
  
\subsection{Data}

The analysis is based on continental U.S. county level data combined with other data resources. The US county level vaccination data "Percent Adults (over 18)  Fully Vaccinated Against COVID-19" (outcome variable) is provided by CDC (September 2021).\footnote{See \cite{Cdc2021-9}.} Other less time sensitive predictor variables such as Percent non-Hispanic Asians, Percent non-Hispanic Black, and  Percent Hispanic are from a  June 2021 CDC dataset.\footnote{See \cite{Cdc2021-6}.}

Although the majority of the U.S. counties have vaccination data available, there are a few exceptions. For example, Texas county-level vaccination data is not present in the CDC dataset. Fortunately, we were able to integrate Texas county level data from the Texas Health and Human Services Agency.\footnote{See \cite{Dshs2021}.} 

Since we believe political affiliation is one of the important factors of how people respond to a public health crisis such as COVID-19, we downloaded the precinct-level U.S. 2020 election results from the New York Times’s github site.\footnote{See New York Times \cite{Nyt2021}.} The original precinct data are in GeoJSON format including “detailed data (...) representing 89 percent of all votes cast” (New York Times \cite{Nyt2021}). The precinct level election data is then aggregated to U.S. county level data based on county’s FIPS code using ArcGIS geoprocessing tools from Esri.

ArcGIS Enrichment tools\footnote{See \cite{Esri2021}.} were used to augment the dataset with Esri projected 2021 Census data including county population data (Total Population and Population over 18 Years Old) as well as with income related data (Percent of Food Stamp Recipients).  Our dataset has 2630 observations (counties).

\newpage
\section{Modelling Impact Factors for Vaccine Hesitation\label{SecAnalysis}}

\subsection{Ordinary Least Square  Model (OLS)\label{SubSecRegression}}

To quantify the impact of various factors on vaccination behavior, we first use a multivariate  OLS model. This ensures that the impacts of the predictor variables are measured simultaneously and consequently that the impact of each variable is controlled for the impact of other variables.   

We consider race/ethnicity impacts measured as the percentage of Asians, African Americans, and Hispanics in each of the counties. Note that White is not considered because it is the residual (together with Other Races/Ethnicities). 

We also included two age variables: the percentage of people in each  county between 20 and 25  years $(Perc_{Young25})$ as well as people older than 65 years $(Perc_{Old65})$ assuming that these two age groups have a different  vaccination behaviour because of the low respectively high risk of serious complications when infected with the COVID-19 virus.

Since the two major political parties in the U.S. have diverging views about the need of vaccination and how vaccinations should be implemented, we added a variable that measures the percentage of votes Republicans received in the last presidential election $(Perc_{Rep})$. Note that a low percentage for Republicans indicates a high percentage for Democrats and vice versa (neglecting  votes for presidential candidates other than Democrats or Republicans).

We also included income effects on vaccination behaviour. After running several models with different variables to correct for income effects (median household income, percent of households below poverty level, and percent of households receiving food stamps), we decided to use the percentage of households receiving food stamps $(Perc_{FoodStamps})$ to control for income effects. Overall, the model results are similar regardless of which variable is used to control for income effects. However, the model based on  $Perc_{FoodStamps}$ provides the best results in terms of prediction quality measured as adjusted $r^2$ as well as significance of the other explanatory variables ($p$ values). This is likely because low  income rather than income in  general influences vaccination rates. The related OLS model is shown below in equation \ref{eq:OlsModel} :
\begin{eqnarray}
Perc_{FullVac}&=&\beta_1 Perc_{FoodSt}+\beta_2 Perc_{Asian}+\beta_3 Perc_{Black} +\beta_4 Perc_{Hisp}\nonumber \\
&+&\beta_5Perc_{Young25}
+\beta_6 Perc_{Old65}
+\beta_7 Perc_{Rep}
+ \beta_0
\label{eq:OlsModel}
\end{eqnarray}

\begin{table}[htbp]
\begin{center}
        \caption{OLS Regression Results}
    \label{TabCountyOlsRegr}

% Table created by stargazer v.5.2.2 by Marek Hlavac, Harvard University. E-mail: hlavac at fas.harvard.edu
% Date and time: Thu, Sep 23, 2021 - 11:23:43 AM
\begin{tabular}{@{\extracolsep{5pt}}lc} 
\\[-1.8ex]\hline 
\hline \\[-1.8ex] 
 & \multicolumn{1}{c}{\textit{Dependent variable:}} \\ 
\cline{2-2} 
\\[-1.8ex] & PercVacFull \\ 
 & FullFoodSt \\ 
\hline \\[-1.8ex] 
 PercFoodSt & $-$0.195$^{***}$ (0.043) \\ 
  PercAsian & 0.109$^{***}$ (0.040) \\ 
  PercHisp & 0.072$^{***}$ (0.013) \\ 
  PercBlack & $-$0.354$^{***}$ (0.019) \\ 
  PercOld65 & 0.114$^{**}$ (0.057) \\ 
  PercYoung25 & $-$0.629$^{***}$ (0.100) \\ 
  PercRep & $-$0.522$^{***}$ (0.015) \\ 
  Constant & 0.945$^{***}$ (0.023) \\ 
 \hline \\[-1.8ex] 
Observations & 2,630 \\ 
R$^{2}$ & 0.498 \\ 
Adjusted R$^{2}$ & 0.497 \\ 
Residual Std. Error & 27.831 (df = 2622) \\ 
F Statistic & 371.924$^{***}$ (df = 7; 2622) \\ 
\hline 
\hline \\[-1.8ex] 
\textit{Note:}  & \multicolumn{1}{r}{$^{*}$p$<$0.1; $^{**}$p$<$0.05; $^{***}$p$<$0.01} \\ 
\end{tabular}

\end{center}
\end{table}

The results of the OLS analysis are displayed in Table \ref{TabCountyOlsRegr}. All variables are significant at the 99\% significance level and have the expected sign. The coefficients show by how much the vaccination rate is expected to increase when one of the variables increases by one percentage point. The variable  $Perc_{Rep}$ has the strongest negative effect with a decrease of the expected vaccination rate about half of the percentage change of republican vote. E.g. if the Republican vote increases by 5\%-points the vaccination rate is expected to decrease by 2.5\% (a factor of 0.5). The variable  $Perc_{Black}$ shows the second strongest negative effect. An increase by one percentage point of  $Perc_{Black}$  is related to a decrease of 0.4 of a percentage point in the expected vaccination rate. The latter shows the need for community outreach to target black neighborhoods. 

The strongest positive impact can be observed for the percentage Asians and for seniors in a county with an about 0.1 percentage point increase of the vaccination rate for a each percent increase of the respective population.  This might confirm that the older population is more willing to get vaccinated because of their higher vulnerability.   Also Asians are generally known for taking greater precautions to avoid getting infected. The variable for young people, who are less in danger for hospitalization or death,  has a significant negative impact.

The impact of $Perc_{FoodSt}$ is negative, which reflects that poverty is an important issue when it comes to vaccination hesitancy. The positive impact of $Perc_{Hisp}$  on the expected vaccination rate is significant but quantitatively smaller than  the impact of the other variables. The relatively small impact of $Perc_{Hisp}$ might be explained by two effects that work in different directions resulting in a relatively small overall impact. Hispanics might have a similar vaccination behavior than African Americans (negative impact on vaccination rate). On the other hand, predominately Hispanic neighborhoods were the ones most badly hit at the beginning of the pandemic (positive impact on vaccination rate).\footnote{This argument will be confirmed when the results of the machine learning model will be interpreted using SHAP values in Section \ref{SecCountyRandFor}.}

The OLS results have to be interpreted with great care. We cannot conclude that an increase of one of the variables in a specific county by one percentage point would lead to a lower vaccination rate of $x$-percentage points in that specific county. The result reported above hold only  for the average of all counties. When implementing machine learning in Section \ref{SecCountyRandFor}, we will derive so called SHAP values for each county and each variable. These SHAP values will reflect the impact for each of the explanatory variable on the expected vaccination rate of a specific county.

In addition, interactions between explanatory variables are neglected in the OLS model. For example, an increase of Republican vote might have a different impact on vaccination rates in a poorer county (more food stamp recipients) than in a relatively richer county. A linear model like the one used here cannot capture these effects.\footnote{It is possible to add interaction terms explicitly as additional variables to an OLS model but this is usually not as effective as using a tree-based machine learning model like Random Forest. The latter also does not require to explicitly define interaction terms. It considers interactions in its dynamically optimized structure. } Therefore, in the next section we will apply a nonlinear Random Forest model to consider these effects.

\subsection{Random Forest Model\label{SecCountyRandFor}}

In this section we use a Random Forest model to predict vaccination rates. The idea behind a Random Forest model is presented in Section \ref{SecCountyRandForIdea} and the prediction results are discussed in Section \ref{SecCountyRandForPrediction}. We also quantify the impacts on the vaccination rate for each of the predictor variables separately for each county by using SHAP values (see Section \ref{SecCountyShap}). 

In order to compare the predictive performance of the machine learning model with the OLS model we use the same data as before, but the observations are randomly assigned to a training dataset (2238 observations) and a testing dataset (392 observations). While the training dataset is used to find the optimal parameters for the OLS model and to optimize the Random Forest model, the testing dataset is never used for any type of optimization and is exclusively reserved to validate the prediction quality of the two models. 

Optimizing both the OLS and the Random Forest model exclusively based on the training data and comparing the models based on the testing data is needed because most machine learning models, including Random Forest, have a tendency to overfit when the model's complexity is high.\footnote{While increasing the number of trees in a Random forest model cannot lead to overlearning, increasing other hyper-parameters, such as the tree depth, can lead to overlearning.} Overfitting occurs when a complex machine  learning model approximates the training dataset almost perfectly, but performs poorly on the testing dataset. This would make a comparison between the two models impossible if the complete dataset was used for training and at the same time for validation.    

\begin{figure}[htbp]
  \centering
    \includegraphics[width=1\textwidth]{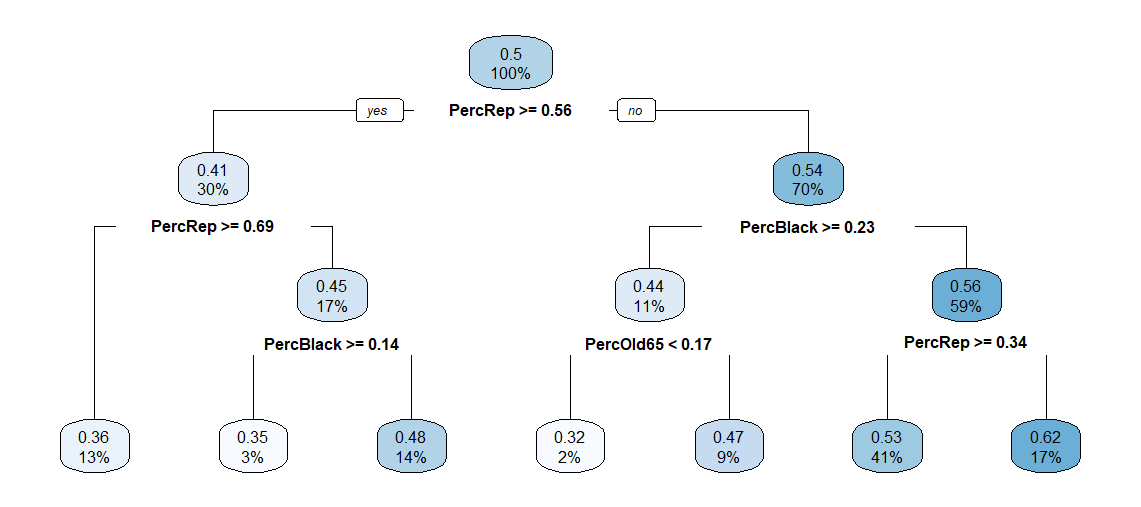}
\caption{Decision Tree}\label{FigDecTree}
\end{figure}
\subsubsection{The Idea Behind a Random Forest Model\label{SecCountyRandForIdea}}
   
A Random Forest model consists of many Decision Tree models (2000 in our case). The prediction of a Random Forest model is calculated as the mean of the predictions from these Decision Tree models. Therefore,  explaining the idea behind a Decision Tree model is a good starting point for understanding  a Random Forest model.

In a Decision Tree (see Figure \ref{FigDecTree} for an example) each observation of a dataset is guided through a treelike structure along internal decision nodes and ends up in one of several bins (terminal nodes). At each of the internal nodes an  observation is tested based on an inequality. Depending on the outcome of the test the observation is moved down to one of two branches (left branch, if test condition is fulfilled, right branch, otherwise). This is repeated at the following internal nodes until the observation reaches the terminal node on the bottom of the tree. The decision rules at each internal node (choosing the predictor variable and the splitting value) are determined by an optimization algorithm based on the training data. 

After decision rules for all nodes in a tree have been established and all training data observations have been sorted into the terminal nodes (bins), the decision tree can be used for predictions. For example a county with  53\% voting for Republicans,  a proportion of 28\% of African Americans, and a proportion of 15\% of people older than 65 moves right at the first node (condition not fulfilled),  moves left at the following node (condition fulfilled), and moves left (condition fulfilled) into the middle terminal node/bin of the tree in Figure \ref{FigDecTree}. 

This county would be predicted to have a vaccination rate of 0.32. The reason is that the average vaccination rate of all training observation (counties) that also ended up in the same terminal node during the training process was 0.32. Overall 45 training observations ended up in the middle terminal node (2\% of the 2,238 training observations).

Decision trees, although intuitive, are called weak predictors because they respond sensitively to small changes in the data or optimization parameters. However, by combining many different decision trees --- different in terms of chosen variables and splitting values at each node --- this problem can be mitigated.  

As stated earlier, a Random Forest prediction\footnote{See \cite{Breiman2001}.} is derived from the mean prediction of all its decision trees and usually delivers better and more stable predictions than the individual decision trees. The idea that a combination of weak predictors can lead to a strong prediction is analogous to the wisdom of crowds phenomenon described in \cite{Galton1907} where visitors at a stock and poultry exhibition in England submitted guesses about the weight of an ox. Although most of the visitors were off with their predictions, surprisingly the mean of all predictions was very close to the real weight of the ox.\footnote{See \cite{Galton1907}.}

In order to generate diverse decision trees, two strategies are employed with Random Forrest: 
\begin{description}
\item  Bagging:\footnote{See \cite{Breiman1996}.}  Each Decision Tree is presented with a different training dataset, which has the same number of records than the original dataset and is generated by  drawing with replacement from the original dataset (Bootstrapping\footnote{See \cite{Efron1979}.}).
\item Random Subspace Method:\footnote{See \cite{Ho1998}.} For each decision tree only a random subset of exogenous variables is considered as candidates for the decision rules. To find  a reasonable value for the number of variables to be considered a rule of thumb suggest to use $\sqrt{M}$ randomly chosen predictor variables, where $M$ is the number of all predictor variables.
\end{description}
\subsubsection{Random Forest Predictions\label{SecCountyRandForPrediction}}
The Ranger R package was used to run the Random Forest model.\footnote{See \cite{RangerPackage} for details about the Ranger package.} The Random Forest model used consists of 2,000 decision trees, with a minimum node size of 5 (default),\footnote{When the minimum node size (observations in a decision node) is reached, a decision tree stops further branching at this node and the node becomes a final node.} and a subset of two variables (rounded down from $\sqrt{7}$; default) that is used as candidate variables for the decision rules. The default values were confirmed by cross validation on the training dataset. 

When comparing the  Random Forest model to the OLS model based on the testing data, the Random Forest model is able to reduce the Mean Absolute Error (MAE) from 8.3\% for the OLS model to 7.8\%  for the Random Forest Model.
\begin{align*}
MAE=\frac{1}{N}\sum_{i=1}^N
\widehat{V_i^{acc}}-V_i^{acc}\quad \mbox{with: $\widehat{V_i^{acc}}:=$ predicted and $V_i^{acc}:=$ true vaccination rate}
\end{align*}
Since $r^2$ was used for both the OLS and the Random Forest model as minimization criteria during the training process, it makes sense to compare the performance also in regards of $r^2$. Based on the testing data the Random Forest Model improved  to $r_{RF}^2=0.441$ compared to $r_{OLS}^2=0.399$ for the OLS. 
\begin{align*}
r^2=\frac{1}{N}\sum_{i=1}^N
\left(\widehat{V_i^{acc}}-V_i^{acc}\right))^2\quad \mbox{with: $\widehat{V_i^{acc}}:=$ predicted and $V_i^{acc}:=$ true vaccination rate}
\end{align*}

Because machine learning models consider interactions between variables and are inherently non-linear they usually provide better predictions compared to OLS models. However, in the past   they were often considered Black-Box models because most machine learning models do not provide information about the quantitative impact of the predictor variables. 

This is no longer true because significant progress has been made on Artificial Intelligence Explanations (AIX) such as LIME and SHAP values.\footnote{While Local Interpretable Model-Agnostic Explanations (LIME) (see \cite{Ribeiro2016}) is based on a local linear approximation of the prediction surface of the underlying machine learning model, SHAP values allow to estimate the impact of each variable for each observation (counties in our case) and for each of the variables separately (see \cite{Lundberg2017}). }  We will use SHAP values in the following section to quantify the impact of each predictor variable in each individual county on the predicted vaccination rate. SHAP values do not only provide information about the why but also about the how much a predictor variable influences the vaccination rate.

\subsubsection{Quantifying the Impact of Predictor Variables Based on SHAP Values\label{SecCountyShap}}

Very recently, SHAP values became popular in the machine learning literature to quantify the impact of predictor variables for a trained machine learning model.\footnote{SHAP values for tree like machine learning models were first introduced by \cite{Lundberg2017} and are now used for a wide range of machine learning models. An introduction to the methodology can be found in \cite{Molnar2020}, Chapter 5 and a more intuitive introduction is provided by \cite{Mazzanti2020Shap}.} We use the R package ShapR to estimate SHAP values for every predictor variable and for each of the 2,630 counties.\footnote{See \cite{ShapRPackage} for details about the package.}
\begin{figure}[htbp]
   \centerline{\includegraphics[width=0.7\textwidth]{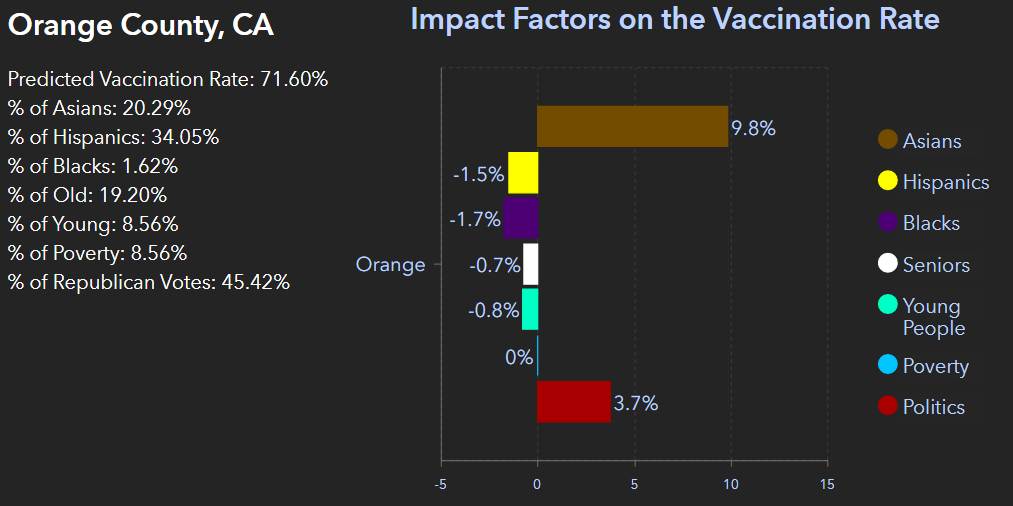}}
   \caption{SHAP Values for Orange County, CA}\label{FigShap3Counties}
\end{figure}

SHAP values are estimated separately for each individual observation (county) and each predictor variable based on a trained (Random Forest) model. In contrast, the coefficients in an OLS model only quantify the impact of the predictor variable on the {\it average} of all observations (counties).

A SHAP value indicates by how much a specific predictor variable contributes (negatively or positively) to a specific county's predicted vaccination rate. SHAP values are not only influenced by the value of the related predictor variable. They also depend on the interaction between all other predictor variables. While OLS coefficients can be estimated analytically, SHAP values are determined numerically based on repeated predicting.  

In Figure \ref{FigShap3Counties} SHAP values together with the empirical values for the related variables are displayed for Orange County, CA. With 20.3\% Asians (see left panel in In Figure \ref{FigShap3Counties} ) the proportion of Asians living in the county is relatively high  compared to the national average (5\%).\footnote{See \cite{Cdc2021-9}. Percent of population older 18, where states provided vaccination data.} This relatively high proportion of Asians combined with the influence from the interaction of the other variables contributes to a 9.8\% increase of the vaccination rate (SHAP value for variable Asians) in Orange County. The low proportion of Hispanics living in Orange county  (4\%) compared to the national average (19\%) together with the interaction among the other variables leads to a SHAP value of -1.5\% (decrease of the predicted vaccination rate). The low proportion of African Americans  living in Orange County (1.6\%) compared to the national average (12\%) impacts the vaccination rate slightly negative with -1.5\%, while counties with a higher proportion of African Americans usually show a higher negative impact. E.g., Figure \ref{FigUniqueBlack} on page \pageref{FigUniqueBlack} shows a decrease of 4.6\% for the proportion of African Americans in Kemper County. The impact of young people, seniors, and poverty is also small in Orange County. In contrast, Politics has a larger impact in Orange County. Figure \ref{FigShap3Counties} shows that Orange county's vote for Republicans was  about 45\% in the recent presidential election, which puts the county's political climate slightly in favor of Democrats. As a consequence the SHAP value for Politics shows a SHAP value (increase of the vaccination rate) of +3.7\%.

All SHAP values are in reference to the national average of the vaccination rate (62.6\%) as of September 21, 2021.\footnote{In order to keep SHAP values, Predicted Vaccination Rates, and the National Vaccination Rate compatible, the National Vaccinations Rate was calculated as a weighted average from the counties' vaccination rates. Weights were based on the population over 18 years of the respective county. Consequently, there are small differences between the National Vaccination Rate reported here and the one reported for the same time by the CDC (65.7\%).} The sum of the National Vaccination Rate and all SHAP values equals the Predicted Vaccination Rate:

\begin{eqnarray*}
\mbox{\bf Orange County, CA: }&& \\
\underbrace{0.716}_{Perc^{Orange}_{PredFullVac}}&=& \underbrace{0.626}_{Perc^{National}_{FullVac}}\\
                                      &+&
                                      \underbrace{0.098}_{SHAP_{Asian}}+
                                      \underbrace{(-0.015)}_{SHAP_{Hisp}} +
                                      \underbrace{(-0.017)}_{SHAP_{Black}}\\
                                      &+&
                                      \underbrace{(-0.007)}_{SHAP_{Senior}}+
                                      \underbrace{(-0.008)}_{SHAP_{Young}}+
                                      \underbrace{0.000}_{SHAP_{Poverty}}+
                                      \underbrace{0.037}_{SHAP_{Politics}}
\end{eqnarray*}

This leaves the more technical question, how is the quantitative impact of a variable $x_i$ for county $i$ (i.e., a predictor variable's SHAP value) predicted? 

SHAP values are generated based  on repeated predictions from the trained Random Forest model.  The basic idea of generating a SHAP value for a specific predictor variable for a specific county $i$ can be illustrated  as follows: 

\begin{description}
\item[Step 1:] The Random Forrest model is used to predict the vaccination rate for all counties based on a dataset where the variable $x$ is switched off. Since setting the values of variable $x$ to zero would introduce bias, the procedure of switching off a variable is embedded in the Decision Tree estimations.
\item[Step 2:] The prediction from Step 1 is repeated but now with the values of variable $x$ switched on.
\item[Step 3:] The (positive or negative) difference for the predicted vaccination rate for each county between Step 2 and Step 1 estimates the contribution of variable $x$ towards the predicted vaccination rate for each of the counties. 
\end{description}

Steps 1 -- 3 are repeated for all possible coalitions of predictor variables, where the simplest one does only consider variable $x$, others consider some predictor variables besides $x$, and the largest coalition considers all predictor variables. The estimated SHAP value for variable $x$ is then calculated as the weighted average of all contributions from all scenarios (coalitions) based on Step 3. The weighting schema ensures that for each county the sum of all SHAP values plus the national mean of all counties vaccination rate equals the predicted vaccination rate. 

\begin{figure}[htbp]
  \centering
    \includegraphics[width=1.0\textwidth]{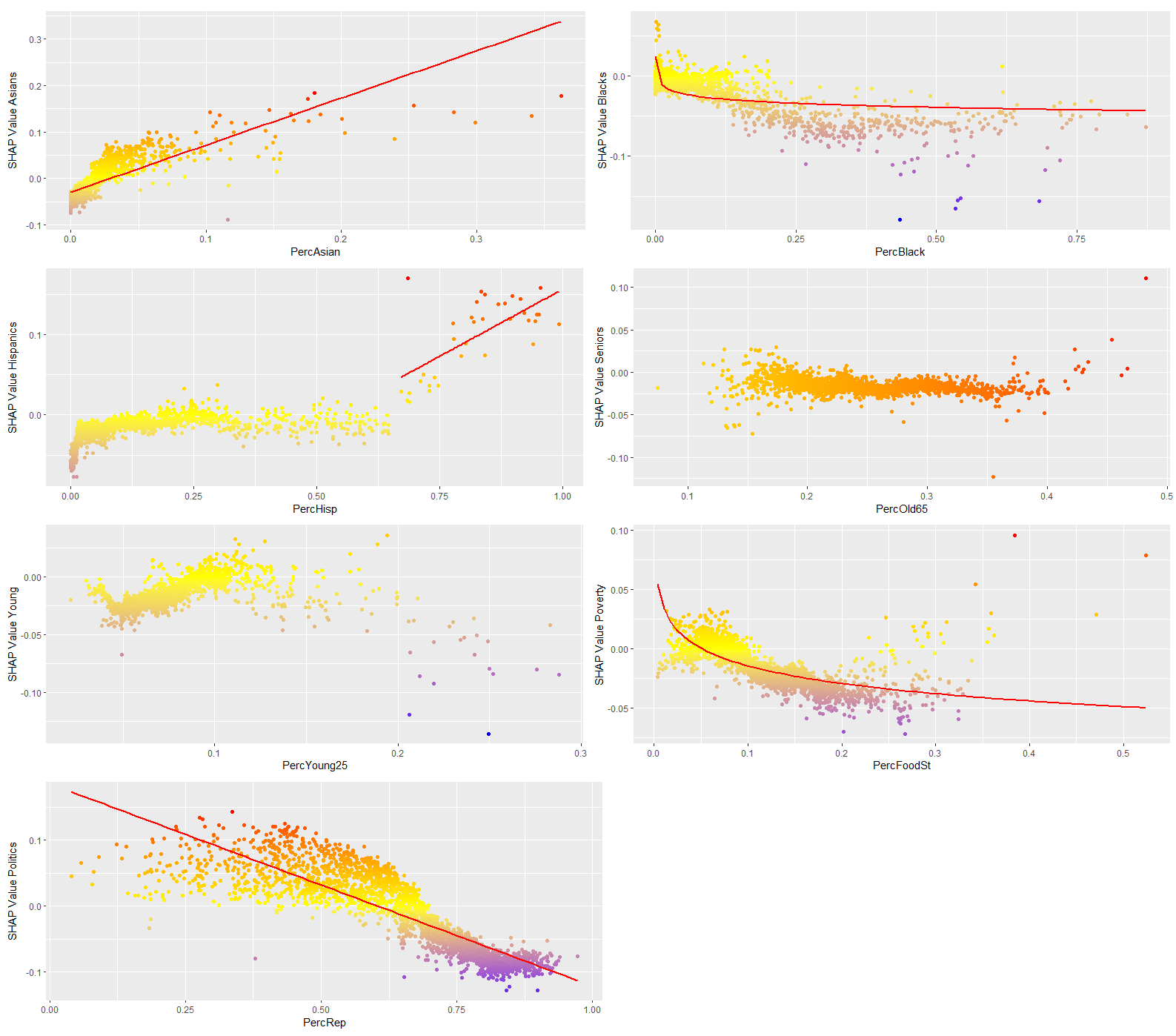}
\caption{SHAP Values for Predictor Variables for Counties from the Continental U.S.}\label{FigShapAllVar}
\end{figure}

The set of all possible coalitions forms what is called in mathematics a power-set.  The number of all possible coalitions in a power-set can be calculated as $2^{NumberPredictors}$, which equals 128 ($2^7$) in our case. Given that we consider 7 predictor variables and 2,630 counties. The algorithm would have to repeat steps 1- 3 approximately 2.4 million times ($128 \cdot 2,630 \cdot 7$). SHAP algorithms like the ShapR algorithm  used here\footnote{See \cite{ShapRPackage}} optimize the procedure and thus lower the amount of iterations. However, the calculation of the SHAP values displayed in Figure \ref{FigShapAllVar} still is very computing intensive. It took about 16 hours on a computer with a 7th generation Intel processor with 8 logical cores to estimate all   SHAP values.

\newpage
\section{Key Findings\label{SecKeyFindings}}

When plotting SHAP values for all analyzed counties organized by predictor variables, trends and non-linearities are present for some of the variables (see Section \ref{SecShapTrends}). 
In addition, by analyzing the SHAP values for a few selected counties unique insights into vacation behavior can be discovered (see Section \ref{SecUniqueInsights}). These insights are a direct benefit of calculating SHAP values, because SHAP values allow to analysis individual observations (counties).
\subsection{SHAP Value Trends by Predictor Variable Based on All Counties\label{SecShapTrends}}
We calculated SHAP values for each of the seven predictor variables for all continental U.S. counties where data were available. They are plotted in the seven diagrams in Figure \ref{FigShapAllVar}. In addition, a map that displays detailed information for each individual county similar to Figure \ref{FigShap3Counties} is available on the Internet.\footnote{See \href{https://www.cpp.edu/~clange/vacmap.html}{https://www.cpp.edu/\texttildelow clange/vacmap.html} for an interactive map.}

Each diagram in Figure \ref{FigShapAllVar} displays data points for 2,630 counties. Each point represents the value for the related predictor variable (horizontal axis)  and the resulting SHAP value (vertical axis) for a specific U.S. county. Negative or positive SHAP values indicate the contribution to the vaccination rate for that county. The red trend lines are just provided for visualization purposes. No claim is made that the SHAP values follow these trends.

The exact impact of the predictor variables varies from county to county, but a large percentage of Republican voters as well as a high percentage of African American residents or a high rate of food stamp recipients impacts the predicted vaccination rate negatively compared to the national average vaccination rate (see panels 7, 2, and 6 in Figure \ref{FigShapAllVar} , respectively) . On the other hand, when a county has a large percentage of Asians, its predicted vaccination rate tends to be higher. 

The impact seem to be non-linear for the predictor variables $PercBlack$ and $PercFoodSt$ and linear for the variables $PercAsian$ and $PercRep$. The impact of age groups and Hispanic population is inconclusive overall, but it seems that there is a positive trend for counties with a predominantly  (greater than 65\%) proportion of Hispanics (see panel 3 in Figure \ref{FigShapAllVar}).  

The results are mostly compatible with the traditional OLS regression analysis (see Section~\ref{eq:OlsModel}). However, the traditional regression analysis also indicates that overall a large percentage of Hispanic population or seniors in a county impact the predicted vacation rate positively, while a high percentage of younger people (between age 20 to 25) tends to negatively affect the vaccination rate. The impact of these variables is mostly inconclusive for the machine learning model.

\subsection{Unique Insights \label{SecUniqueInsights}}
When analysing and comparing selected counties some interesting and unique insights about vaccination hesitancy in different US counties can be discovered. Below are a few examples:

\paragraph{Predominately Hispanic Counties Have High Vaccination Rates:}
\begin{figure}[htbp]
  \centering
  \includegraphics[width=0.57\textwidth]{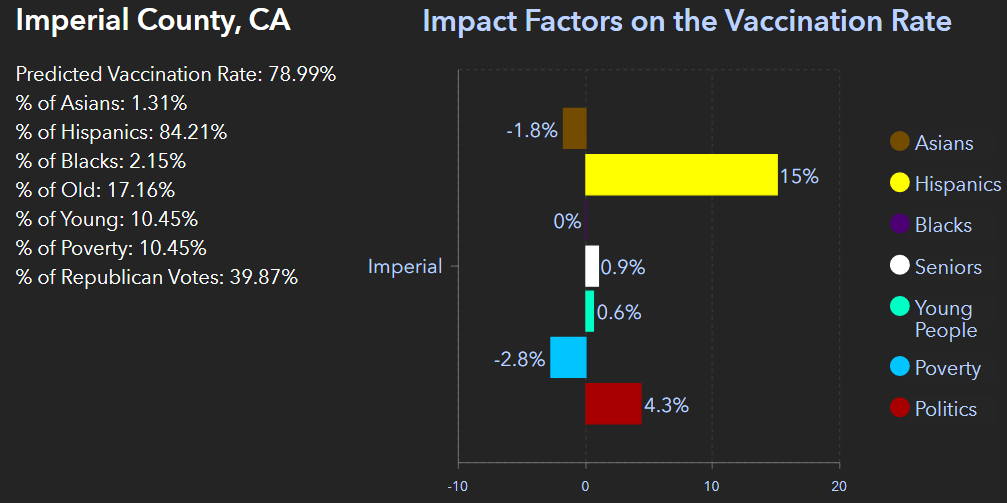}
\caption{Hispanic Vaccine Impact (SHAP) in Imperial County, CA}\label{FigUniqueHispanicic}
\end{figure}
Although we could not derive a clear overall trend for the impact of the  predictor $PercHisp$ in Figure \ref{FigShapAllVar} on page \pageref{FigShapAllVar}, there is a clear and strong positive trend for counties with a predominately  Hispanics population. The higher the population of Hispanics the higher is the vaccination rate as long as the Hispanic population share exceeds 65 percent. A possible reason for this finding could be that in the first wave of COVID counties with a high Hispanic population share were hit hard with hospitalizations and death. Experiencing this first hand might have lead to high acceptance of vaccinations. This is supported by Figure \ref{FigUniqueHispanicic}  showing SHAP values for  Imperial County, CA. This county has 84 percent of Hispanic residents and is also high in poverty (10\% receiving food stamps). However, it has one of the highest vaccination rates in the country and the largest impact on the vaccination rate (+15\%) stems from the SHAP value for Hispanic ethnicity. 

\begin{figure}[htbp]
  \centerline{
  \includegraphics[width=0.565\textwidth]{Orange.png}
    \includegraphics[width=0.575\textwidth]{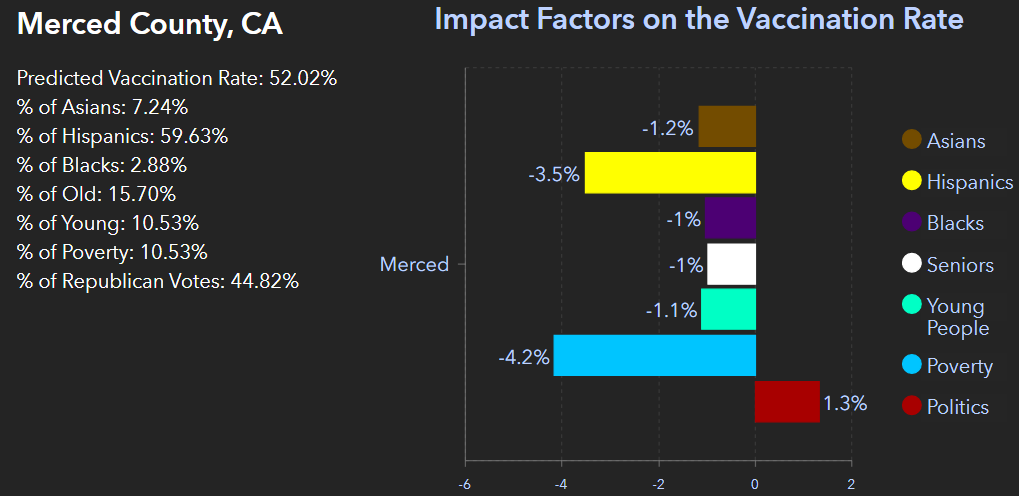}}
\caption{Different Political Impact in the Counties of Orange, CA and Merced, CA}\label{FigShapOcAndMerced}
\end{figure}
\paragraph{Politics Matters but the Impact Varies Even in Similar Political Climates:}
When analyzing the impact of political affiliation many papers consider presidential votes as the only predictor variable. This neglects controlling for the effects of other variables like income or race and it also neglects the interactions between predictor variables. For example, survey results published by the Kaiser Family Foundation (see \cite{Kaiser2021}) indicate that Republicans are much more likely to say no to vaccinations, but the only predictor variable that was used to analyze this was the presidential vote for Republicans. 

One of the key benefits of this study is to consider interaction effects of multiple variables. Why this is important can be seen in Figure \ref{FigShapOcAndMerced}: Both Orange County, CA and Merced County, CA have about 45 percent Republican voters, resulting in a political climate that is slightly Democratic. Although the proportions of the presidential votes are almost identical, the impact of Politics on the vaccination rate is more than double in Orange County compared to Merced County (+3.7\% vs. +1.3\%). This might be attributed to an indirect effect of the  large Asian population in Orange County (20 percent Asian population), which might have increased the effect of the variable Politics in Orange County. This is in addition to the direct effect Asians have on the vaccination rate which is +9.8\% in Orange County compared to -1.3\% in Merced.

\paragraph{Not all Asian Communities are Equal:}
\begin{figure}[htbp]
  \centerline{
  \includegraphics[width=0.57\textwidth]{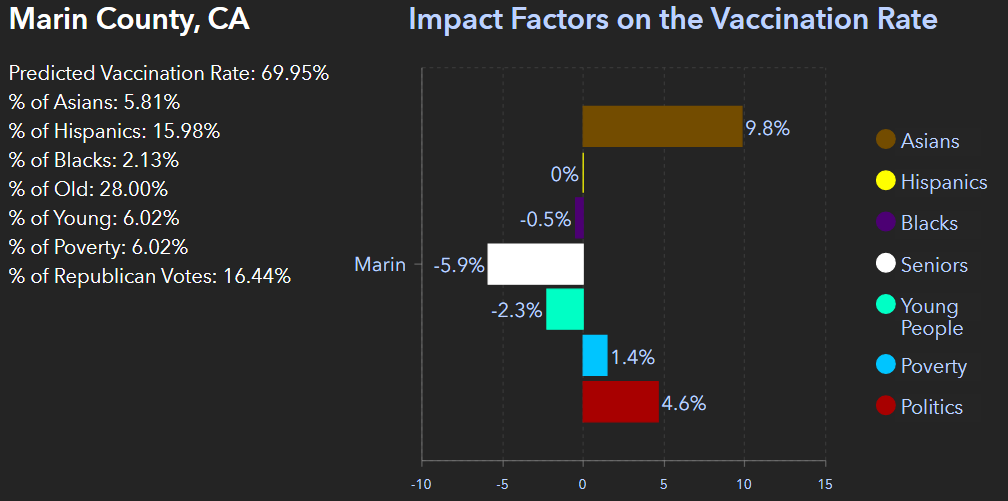}
    \includegraphics[width=0.57\textwidth]{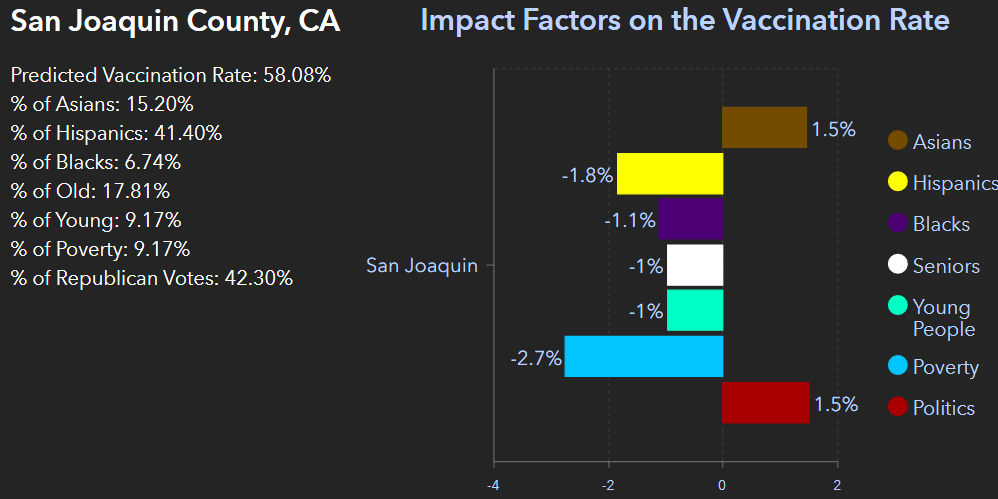}}
\caption{ASIAN Impact on the Vaccination Rate (SHAP) in Marin, CA and San Joaquin Counties, CA }\label{FigUniqueAsians}
\end{figure}

The model reveals for most counties that when a county has a larger percentage of Asians, its predicted vaccination rate tends to be higher, which might be due to cultural characteristics and vaccine mobilization efforts in local Asian communities (see Figure \ref{FigShapAllVar} on page \pageref{FigShapAllVar}). But not all Asian communities are equal. In the example in Figure \ref{FigUniqueAsians}, Marin County, CA has a merely 5.8 percent Asian population, while the nearby San Joaquin County has 15 percent Asian population. However, due to other disparities in the two Asian communities, surprisingly the Asian population influence is higher on the vaccination rate in Marin County than in San Joaquin County (+9.8\% vs. +1.5\%). One of these disparities could be that many Asians in Marin County work in the computer industry. In contrast, in San Joaquin County many Asians work in the farming industry.

\paragraph{African American Communities Vary across Different Geographies:}
\begin{figure}[htbp]
  \centerline{
  \includegraphics[width=.57\textwidth]{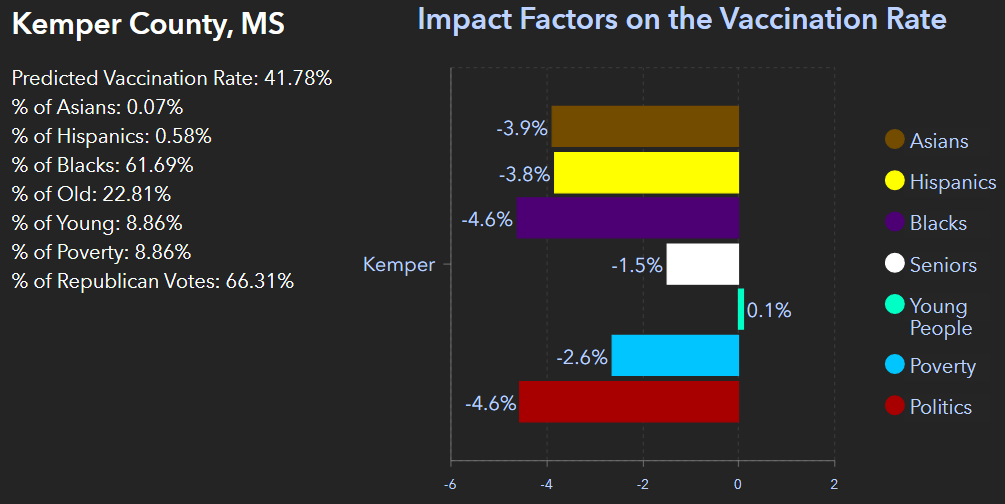}
    \includegraphics[width=0.57\textwidth]{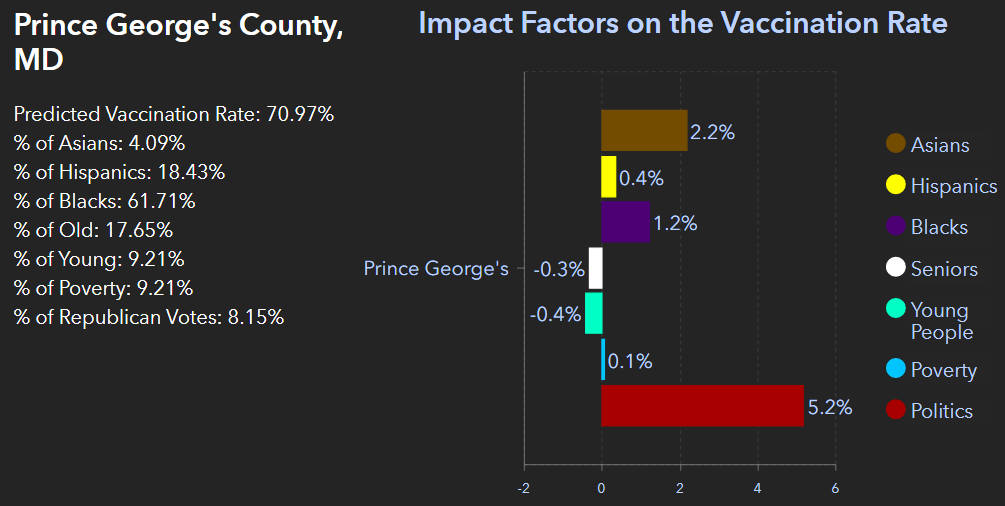}}
\caption{Comparing the Kemper, MS and PrinceGeorge, MD}\label{FigUniqueBlack}
\end{figure}

Although the impact of the variable $PercentBlack$ on the vaccination rate is negative for most counties (see Figure \ref{FigShapAllVar} on page \pageref{FigShapAllVar}), this is not true in all cases. Figure \ref{FigUniqueBlack} shows proportions and SHAP values for Prince George County, MD and Kemper County, MS. Both counties are about 62\% African American. In Kemper County, the SHAP value for Black reflects a negative impact of 4.6\% on the vaccination rate. In contrast, in Prince George County, the impact is 3.4\% positive. Because Prince George County borders Washington DC, it is possible that residents have a very different exposure and understanding of current affairs than the residents in Kemper County and are thus more likely to get vaccinated.

\section{Summary}
The study uses an OLS model as benchmark and a Random Forest model to estimate which socioeconomic factors impact the vaccination rates in U.S. continental counties. 

The Random Forest model generates a higher prediction quality than the OLS model. More importantly, the Random Forest model considers interactions between the predictor variables, accounts for non-linearities, and provides predictions for each individual county's vaccination rate. By implementing an Artificial Intelligence Explanations algorithm (SHAP values), it is possible in every county to quantify how much each impact factor contributes to the county's vaccination rate. 

For most counties a higher percentage vote for Republicans, a greater African American population share, and a higher poverty rate lower the vaccination rate. While  a higher Asian population share increases the predicted vaccination rate.

The impact on the vaccination rate from the Hispanic population proportion, the percentage of seniors, and the one for young people in a county has either a weak or inconclusive impact. However, when the population share of Hispanics increases 65\% the impact of Hispanics starts to increase.

An interactive online mapping dashboard that identifies impact factors for individual U.S. counties is available at \href{https://www.cpp.edu/~clange/vacmap.html}{https://www.cpp.edu/\texttildelow clange/vacmap.html}. It is apparent that the influence of impact factors is not universally the same across different geographies.
\newpage
\printbibliography
\end{document}